\documentclass[10pt, a4paper]{article}
\usepackage{lrec2022} 
\usepackage{multibib}
\newcites{languageresource}{Language Resources}
\usepackage{graphicx}
\usepackage{tabularx}
\usepackage{todonotes}
\usepackage{placeins}
\usepackage{float}
\usepackage{soul}
\usepackage{titlesec}
\titleformat{\section}{\normalfont\large\bfseries\center}{\thesection.}{1em}{}
\titleformat{\subsection}{\normalfont\SmallTitleFont\bfseries\raggedright}{\thesubsection.}{1em}{}
\titleformat{\subsubsection}{\normalfont\normalsize\bfseries\raggedright}{\thesubsubsection.}{1em}{}
\renewcommand\thesection{\arabic{section}}
\renewcommand\thesubsection{\thesection.\arabic{subsection}}
\renewcommand\thesubsubsection{\thesubsection.\arabic{subsubsection}}

\usepackage{epstopdf}
\usepackage[utf8]{inputenc}

\usepackage{hyperref}
\usepackage{xstring}
\usepackage{multirow}
\usepackage{color}

\title{RoBERTuito: a pre-trained language model for social media text in Spanish}

\name{Juan Manuel Pérez$^{12}$, Damián A. Furman$^{12}$, Laura Alonso Alemany$^{3}$, Franco Luque$^{23}$}

\address{$^1$Universidad de Buenos Aires, $^2$CONICET, $^3$Universidad Nacional de Córdoba \\
         \{jmperez, dfurman\}@dc.uba.ar, \{francolq, alemany\}@famaf.unc.edu.ar \\
         }

\abstract{
    Since BERT appeared, Transformer language models and transfer learning have become state-of-the-art for natural language processing tasks. Recently, some works geared towards pre-training specially-crafted models for particular domains, such as scientific papers, medical documents, user-generated texts, among others. These domain-specific models have been shown to improve performance significantly in most tasks; however, for languages other than English, such models are not widely available.
    In this work, we present \robertuito{}, a pre-trained language model for user-generated text in Spanish, trained on over 500 million tweets. Experiments on a benchmark of tasks involving user-generated text showed that \robertuito{} outperformed other pre-trained language models in Spanish. In addition to this, our model has some cross-lingual abilities, achieving top results for English-Spanish tasks of the Linguistic Code-Switching Evaluation benchmark (LinCE) and also competitive performance against monolingual models in English Twitter tasks. To facilitate further research, we make \robertuito{} publicly available at the HuggingFace model hub together with the dataset used to pre-train it.
     \\ \newline 
}

\begin{document}

\newcommand{\robertuito}[0]{RoBERTuito}
\newcommand{\robertuitounc}[0]{\robertuito{}$_{uncased}$}
\newcommand{\robertuitocas}[0]{\robertuito{}$_{cased}$}
\newcommand{\robertuitodea}[0]{\robertuito{}$_{deacc}$}

\newcommand{\mbf}[1]{\mathbf{#1}}

\newcommand{\bertweet}[0]{BERTweet}
\newcommand{\bert}[0]{BERT}
\newcommand{\beto}[0]{BETO}
\newcommand{\mbert}[0]{mBERT}
\newcommand{\roberta}[0]{RoBERTa}
\newcommand{\xlm}[0]{XLM-R}
\newcommand{\xlmbase}[0]{XLM-R$_{BASE}$}
\newcommand{\xlmlarge}[0]{XLM-R$_{LARGE}$}

\newcommand{\lince}[0]{LinCE}

\maketitleabstract

\section{Introduction}

Pre-trained language models have become a basic building block in the area of natural language processing. In the last years, since the introduction of the transformer architecture \cite{vaswani2017attention}, they have been used in many other different natural language understanding tasks, outperforming previous models based on recurrent neural networks. BERT, GPT-2, and RoBERTa are some of the most well-known such tools.

Most models are trained on large-scale corpora taken from news or Wikipedia, which are considered general enough to comprise a large part of the language. Some domains, however, are very specific and have their own vocabulary, jargon, or complex expressions. The medical or scientific domains, for example, use terms and concepts which are not found in a general corpus or occur just a few times. In some other cases, words have specific meanings within a particular domain. Colloquial language --as found in Twitter and other social networks-- is more informal, with slang and other expressions which rarely occur in Wikipedia.

For these reasons, a number of pre-trained models have been created to handle these domains. SciBERT \cite{beltagy-etal-2019-scibert} and MedBERT \cite{rasmy2021med} are examples of domain-specific models. For user-generated text, many models have been trained on Twitter for different languages. However, Spanish lacks pre-trained models for user-generated text, or they are not easily available in the most popular model repositories, such as the HuggingFace model hub. This hinders the development of accurate applications for user-generated text in Spanish.

In this paper, we present \robertuito{}, a large-scale transformer model for user-generated text trained on Spanish tweets. We show that \robertuito{} outperforms other Spanish pre-trained models for a number of classification tasks on Twitter. In addition to this, and due to the collection process of our pre-training data, \robertuito{} is a very competitive model in multilingual and code-switching settings including Spanish and English. Our contributions are the following:

\begin{itemize}
    \item We publish the data used to train \robertuito{} (around 500M tweets in Spanish), to facilitate the development of other language models or embeddings, also using subsets of the corpus that model specific subdomains, like regional or thematic variants.
    \item We make the weights of our model available through the HuggingFace model hub, thus sparing computation for researchers with no access to the computational power or simply sparing extra computation, while making the model transparent (albeit not interpretable).
    \item We set up a benchmark for classification tasks involving user-generated text in Spanish.
    \item We assess the performance of domain-specific models with respect to general-language models, showing that the first outperform the latter in the corresponding domain-specific tasks.
    \item We assess the impact of preprocessing strategies for our models: cased input vs. uncased input text vs. uncased input text without accents, showing that the uncased version of the corpus yields better performance.
    \item We also evaluate our model in a code-switching benchmark for Spanish-English and in a small number of English tasks, both for user-generated text, showing that it achieves competitive results.
\end{itemize}

\section{Previous Work}
Language models based on transformers \cite{vaswani2017attention} have become a key component in state-of-the-art NLP tasks, from text classification to natural language generation. One of the most popular transformer-based tools, \bert{} \citelanguageresource{devlin-etal-2019-bert} is a neural bidirectional language model trained on the Masked-Language-Model (MLM) task and in the Next-Sentence-Prediction (NSP) task. This language model can then be fine-tuned for a downstream task or can be used to compute contextualized word representations. \roberta{} \citelanguageresource{liu2019roberta} is an optimized pre-training approach that differs from BERT in four aspects: it trains the model with more data; it removes the NSP objective; it uses longer batches; and it dynamically changes the masking pattern applied to the data. These models, along with GPT \citelanguageresource{radford2018improving}, supposed breakthroughs in the performance on benchmarks such as GLUE \cite{wang-etal-2018-glue}. \newcite{nozza2020mask} provides a good overview of the \bert{}-based language models.

After the explosion of language models based on transformers, some models have been trained on corpora that target more specifically a domain of interest instead of generic texts such as Wikipedia or news. For example, SciBERT \citelanguageresource{beltagy-etal-2019-scibert} is a \bert{} model trained on scientific texts, and MediBERT \citelanguageresource{rasmy2021med} was crafted on medical documents. AlBERTo \cite{polignano2019alberto} is one of the first models trained on tweets --particularly, in Italian. \bertweet{} \citelanguageresource{bertweet} is a \roberta{} model trained on about 850M tweets in English, a part of them related to the COVID-19 pandemic.

Multilingual models have also been successful at many tasks comprising more than one language. Multilingual \bert{} (\mbert{}) \cite{devlin-etal-2019-bert} was pre-trained on the concatenation of the top-104 languages from Wikipedia. In a parallel fashion, \xlm{} \citelanguageresource{conneau-2020-xlm} was trained using \roberta{} guidelines on the concatenation of Common Crawl data containing more than 100 languages, obtaining a considerable performance boost over several multilingual tasks while keeping competitive with monolingual models.

\beto{} \cite{canete2020spanish} was the first publicly available pre-trained model in Spanish, following mainly a BERT style of training (MLM + NSP) with some ideas taken from \newcite{liu2019roberta}. More recently, some other pre-trained models have been developed for this language, such as \roberta{}-BNE \cite{gutierrezfandino2021spanish}, a RoBERTa-based model trained on a database of 500GB of all the \emph{.es} domain websites. BERTin \citelanguageresource{BERTIN} is also a RoBERTA-based model, for whose development the authors explored sampling strategies over the Spanish portion of the \emph{mc4} corpus \cite{JMLR:v21:20-074}.

To the best of our knowledge, TwilBERT \citelanguageresource{gonzalez2021twilbert} is the only specialized pre-trained model on Twitter data for the Spanish language. However, this model has some limitations: first, the training data is not available, making it not auditable. Second, it is not clear how long its pre-training was. Third, the authors used a variant of the NSP task adapted to Twitter (Reply Order Prediction), in spite of many works showing that the type of training based on \roberta{} (MLM only) improves performance on downstream tasks. Finally, the model is not easily available (for instance, in the HuggingFace model hub\footnote{\url{https://huggingface.co/models}}), which makes its use difficult.

\section{Data}

In this section, we describe the tweet collection process used to build the corpus to train \robertuito{}. Twitter's free access streaming API (also known as \emph{Spritzer}) provides a sample of around 1\% of the overall published tweets, supposedly random, although some studies have shown some concerns about the possible manipulation of this sample \cite{pfeffer2018tampering}. Unrepresentative, biased samples may produce biased behaviours in the resulting model and systematic, possibly harmful errors in downstream tasks that use this model. That is why we make available the training dataset and the specifics of the model, so that it can be fully inspected in case biases are suspected. In following releases of this tool, an extensive audit of the training corpus will be carried out.

First, we downloaded a Spritzer collection uploaded to Archive.org dating from May 2019 \footnote {\url {https://archive.org/details/archiveteam-twitter-stream-2019-05}}. From this, we only keep tweets with \emph{language} metadata equal to Spanish, and mark the users who posted these messages. Then, the tweetline from each of these marked users was downloaded. We decided to download data from users already represented in the initial collection to facilitate user-based studies in this dataset, and also because we believe the original sample of users to be representative, and thus we hope to maintain this representativeness by sampling from the same users. In total, the collection consists of around 622M tweets from about 432K users.

Finally, we filtered tweets with less than six tokens, because language contained in those is very different from the language in longer tweets. To identify tokens we used the tokenizer trained in BETO \citelanguageresource{canete2020spanish}, without counting character repetitions and emojis. This leaves a training corpus of around 500M tweets, which we split in many files to facilitate reading in later processes. The code for the collection process can be found at \url {https://github.com/finiteautomata/spritzer-tweets}.

Something to remark is that this collection process allows the data to contain code-mixed text or even tweets from other languages, as we only required the post on the original sample to be in Spanish. While other works such as \newcite{bertweet} required every tweet to be in English, we let other languages to be included in the pre-training data. A rough estimate of the language population using \emph{fasttext}'s language-detection module \cite{joulin2016bag} suggests that 92\% of the data is in Spanish, 4\% in English, 3\% in Portuguese, and the rest in other languages.

\section{RoBERTuito}
In this section, we describe the training process of \robertuito{}. Three versions of \robertuito{} were trained: a \emph{cased} version which preserves the case found in the original tweets, an \emph{uncased} version, and a \emph{deacc} version, which lower-cases and removes accents on tweets. Normative Spanish prescribes marks for (some) accents in words, but their usage is inconsistent in user-generated text, so we want to test if removing them improves the performance on downstream tasks.

For each of the three configurations, we trained tokenizers using \emph{SentencePiece} \cite{kudo-richardson-2018-sentencepiece} on the collected tweets, limiting vocabularies to 30,000 tokens. We used the \emph{tokenizers} library \cite{hftokenizers2019} which provides fast implementations in Rust for many tokenization algorithms.


\subsection{Preprocessing}

Preprocessing is key for models consuming Twitter data, which is quite noisy, have user handles (@username), hashtags, emojis, misspellings, and other non-canonical text. \newcite{bertweet} tried two normalization strategies: a soft one, in which only minor changes are performed to the tweet such as replacing usernames and hashtags, and a more aggressive one using the ideas of \newcite{han2011lexical}. The authors found no significant improvement by using the harder normalization strategy. Having this in mind, we followed an approach similar to the one used both in this work and in \newcite{polignano2019alberto}:

\begin{itemize}
    \item Character repetitions are limited to a max of three
    \item User handles are converted to a special token \verb|@usuario|
    \item Hashtags are replaced by a special token \verb|hashtag| followed by the hashtag text and split into words if this is possible 
    \item Emojis are replaced by their text representation using \emph{emoji} library\footnote{\url{https://github.com/carpedm20/emoji/}}, surrounded by a special token \verb|emoji|.
\end{itemize}

\subsection{Architecture and training}

A \roberta{} base architecture was used in \robertuito{}, with 12 self-attention layers, 12 attention heads, and hidden size equal to 768, in the same fashion as BERTweet \cite{bertweet}. We used a masked-language objective, disregarding the next-sentence prediction task used in BERT or other tweet-order tasks such as those used in \newcite{gonzalez2021twilbert}.

Taking into account successful hyperparameters from \roberta{} and \bertweet{}, we decided to use a large batch size for our training. While an 8K batch size is recommended in \roberta{}, due to resource limitations, we decided to balance the number of updates using a 4K size. To check convergence, we first trained an uncased model for 200K steps. After this, we then proceeded to run it for 600K steps for the three models. This is roughly half the number of updates used to train \beto{} (and also \bertweet{}) but this difference is compensated by the larger batch size used to train \robertuito{}. 

We trained our models for about three weeks on a v3-8 TPU and a preemptible \emph{e2-standard-16} machine on GCP. Our codebase uses \emph{HuggingFace's transformers} library and their \roberta{} implementation \cite{wolf-etal-2020-transformers}. Each sentence is tokenized and masked dynamically with a probability equal to $0.15$. Further details on hyperparameters and training can be found in the Appendix \ref{app:hyperparameters}.

\section{Evaluation}

\begin{table*}[ht!]
    \centering
    \begin{tabular}{c lll r}
        Language                          & Tasks               & Type of task                          & Dataset               & Num. posts  \\
        \hline
        \multirow{4}{*}{Spanish}          & Sentiment analysis    & \multirow{4}{*}{Text Classification}  & TASS 2020 Task A      & $14,500$      \\
                                          & Emotion analysis      &                                       & TASS 2020 Task B      &  $8,400$       \\
                                          & Hate speech detection &                                       & HatEval               & $6,600$       \\
                                          & Irony detection       &                                       & IrosVA 2019           & $9,000$       \\
        \hline \rule{0pt}{1.2em}
        \multirow{3}{*}{English}          & Sentiment analysis    & \multirow{3}{*}{Text Classification}  & SemEval 2017 Task 4   & $61,900$       \\
                                          & Emotion analysis      &                                       & TASS 2020 Task B      & $7,303$        \\
                                          & Hate speech detection &                                       & HatEval               & $13,000$       \\
        \hline \rule{0pt}{1.2em}
        \multirow{3}{*}{Spanish-English}  & Sentiment analysis    & Text Classification                   & \multirow{3}{*}{LinCE}& $18,789$      \\
                                          & POS tagging           & Text Labelling                        &                       & $42,911$       \\
                                          & NER                   & Text Labelling                        &                       & $67,233$       \\
        \hline

    \end{tabular}
    \caption{Evaluation tasks for \robertuito{}. Tasks are grouped by setting: Spanish-only tasks, English-only tasks, and code-mixed Spanish-English tasks. Num. posts is the number of instances contained in the dataset of each task.}
    \label{tab:evaluation_settings}
\end{table*}

We evaluated \robertuito{} in two monolingual settings (Spanish and English) and also in a code-mixed benchmark for tweets containing both Spanish and English. As the collection process allowed non-Spanish tweets to be included, we assess not only the performance of our model in Spanish, but also in other environments. Table \ref{tab:evaluation_settings} summarizes the evaluation tasks.

\subsection{Spanish evaluation}

For the \textbf{Spanish} evaluation, we set a benchmark for this model following TwilBERT \cite{gonzalez2021twilbert}, AlBERTo \cite{polignano2019alberto} and BERTweet \cite{bertweet}. Four classification tasks for Twitter data in Spanish were selected, three of them coming from the Iberian Languages Evaluation Forum (IberLEF) and one from SemEval 2019: \textbf{sentiment analysis}, \textbf{emotion analysis}, \textbf{irony detection} and \textbf{hate speech detection}. These tasks are particularly relevant in the context of social media analysis and provide a broad perspective about how our model is able to capture useful information on this specific domain.

Regarding sentiment analysis, we relied on the TASS 2020 \cite{tass2020garcia} dataset. This dataset uses a three-class polarity model (negative, neutral, positive) and is separated according to different geographic variants of Spanish. For the purpose of benchmarking our model, we disregarded these distinctions and merged all the data into a single dataset, with respective train, dev, and test splits.

For emotion analysis, we also used the dataset from the TASS 2020 workshop, \emph{EmoEvent} \cite{plaza-del-arco-etal-2020-emoevent}. This is a multilingual emotion dataset labeled with the six Ekman's basic emotions (\emph{anger}, \emph{disgust}, \emph{fear}, \emph{joy}, \emph{sadness}, \emph{surprise}) \cite{ekman1992emotions} plus a neutral emotion. It was built retrieving tweets associated with eight different global events from different domains (political, entertainment, catastrophes or incidents, global commemorations, etc.), so emotions are always related to a particular phenomenon. We only kept the Spanish portion of it, containing 8,409 tweets.

Hate speech detection in social media has gained much interest in the past years, based on the need to act against the spread of hate messages that develops in parallel with an increasing amount of user-generated content. It is a difficult task that requires a deep, contextualized understanding of the semantic meaning of a tweet as a whole. For this reason, we selected the \emph{hatEval} Task A dataset \cite{hateval2019semeval}, which is a binary classification task for misogyny and racism, to benchmark our model. The authors collected the dataset by three combined strategies: monitoring potential victims of hate accounts, downloading the history of identified producers of hateful content, and filtering Twitter streams with keywords. This dataset distinguishes between hate speech targeted to individuals and generic hate speech, and between aggressive and non-aggressive messages. For this work, we do not consider these classifications, and we are interested in predicting only the binary label of whether the tweet is hateful or not. The Spanish subset of this dataset comprises 6,600 instances.

Irony detection has also recently gained popularity. Many works show that it has important implications in other natural language processing tasks that require semantic processing. \newcite{gupta-yang-2017-crystalnest} showed that using features derived from sarcasm detection improves the performance on sentiment analysis. In addition to this, user-generated content is a rich and vast source of irony, so being able to detect it is of particular importance for the domain of social networks. IroSVa \cite{ortega2019overview} is a recent dataset published in 2019, that has the particularity of considering the messages not as isolated texts but with a given context (a headline or a topic). It consists of 7,200 train and 1,800 test examples divided into three geographic variants from Cuba, Spain, and Mexico, each with a binary label indicating if the comment contains irony or not. Unlike the previous three tasks mentioned here, this dataset contains not only messages from Twitter but also news comments and debate forums as 4forums.com and Reddit.

We compare \robertuito{} performance for these tasks with other Spanish pre-trained models: \beto{} \citelanguageresource{canete2020spanish}, RoBERTa-BNE \citelanguageresource{gutierrezfandino2021spanish} and BERTin \citelanguageresource{BERTIN}. All these models share a base architecture with a similar number of parameters to our model.

\subsection{English evaluation}

As for \textbf{English}, we tested \robertuito{} in three tasks: emotion analysis, hate speech detection, and sentiment analysis. For emotion analysis and hate speech we used the English sections from the aforementioned datasets (\emph{EmoEvent} and \emph{HatEval}), while for sentiment analysis \emph{SemEval 2017 Task-4} dataset \cite{rosenthal-2017-semeval} was used, which shares the same labels as its Spanish counterpart (negative, neutral, positive).

In this case, we compare \robertuito{} abilities in English with monolingual models: \bert{} \citelanguageresource{devlin-etal-2019-bert}, \roberta{} \cite{liu2019roberta}, and \bertweet{} \citelanguageresource{bertweet}; and also against multilingual models such as \xlmbase{} \cite{conneau-2020-xlm} and \mbert{}. While all these models share a base architecture, the different vocabulary sizes and number of parameters (see Appendix \ref{app:vocabulary_efficiency}) make the comparison a bit more difficult.

\subsection{Code-switching evaluation}
Finally, we assessed the code-switching abilities of our model in the \emph{Linguistic Code-Switching Evaluation Benchmark} (\lince{}) \citelanguageresource{aguilar-etal-2020-lince}. \lince{} comprises five tasks for code-switched data in several language pairs (Spanish-English, Hindi-English, Modern Standard Arabic-Egyptian Arabic, Arabic-English, among others), many of which were part of previous shared tasks. We evaluated \robertuito{} on three different tasks of the benchmark: part of speech (POS) tagging \citelanguageresource{alghamdi-etal-2016-part}, named entity recognition (NER), and sentiment analysis \citelanguageresource{patwa2020sentimix}. As the collection process of the data centered on Spanish-speaking users, some of which also speak English and Spanglish \footnote{The morphosyntactic mixture of Spanish and English}, we test \robertuito{} on the Spanish-English subsection of the benchmark.

This benchmark has a centralized evaluation system, not releasing gold labels for the test split of the tasks. We evaluated our models in the dev datasets and compared our results against the ones provided by \newcite{winata-etal-2021-multilingual}, which achieves the best performance for the NER and POS tagging. As competing models to \robertuito{} for the Spanish-English evaluation, we have \mbert{}, \xlm{} (both in base and large architecture) and monolingual models \bert{} and \beto{}.

\subsection{Model fine-tuning}

We followed fairly standard practices for fine-tuning, most of which are described in \newcite{devlin-etal-2019-bert}. For sentence classification tasks, we fine-tuned the pre-trained models for 5 epochs with a triangular learning rate of $5 10^{-5}$ and a warm-up of 10\% of the training steps. The best checkpoint at the end of each epoch was selected based on each task metric.

For sentence classification tasks, a linear classifier is put on top of the representation of the \verb|[CLS]| token. For token classification tasks (NER and POS tagging), we predicted the word tag by putting a linear classifier on top of the first corresponding sub-word representation.

\section{Results}
\begin{table*}[t!]
    \centering
    \begin{tabular}{lccccr}
        \hline
        Model                 &   Hate   & Sentiment          &    Emotion       &   Irony         &  Score \\
        \hline
        \robertuitounc{}    &  $\mbf{80.1}$ & $\mbf{70.7}$ & $\mbf{55.1}$ & $73.6$ & $\mbf{69.9}$ \\
        \robertuitodea{}    &  $79.8$ & $70.2$ & $54.3$ & $\mbf{74.0}$ & $69.6$ \\
        \robertuitocas{}    &  $79.0$ & $70.1$ & $51.9$ & $71.9$ & $68.2$ \\
        \roberta{}          &  $76.6$ & $66.9$ & $53.3$ & $72.3$ & $67.3$ \\
        BERTin              &  $76.7$ & $66.5$ & $51.8$ & $71.6$ & $66.7$ \\
        BETO$_{cased}$      &  $76.8$ & $66.5$ & $52.1$ & $70.6$ & $66.5$ \\
        BETO$_{uncased}$    &  $75.7$ & $64.9$ & $52.1$ & $70.2$ & $65.7$ \\
        \hline
    \end{tabular}
    \caption{Evaluation results for Spanish classification tasks: hate speech detection, sentiment analysis, emotion analysis and irony detection. Results are expressed as the mean Macro F1 score of 10 runs of the classification experiments. Bold indicates best performing models}
    \label{tab:evaluation_results}
\end{table*}

Table \ref{tab:evaluation_results} displays the results for the evaluation for the four proposed classification tasks in Spanish. Figures are expressed as the mean of 10 runs of the experiments, along with a \textbf{score} averaging the metrics for each task in a similar way as the GLUE score. We can observe that, in most cases, all \robertuito{} configurations perform above other models, in particular for hate speech detection and sentiment analysis. For most tasks, no big differences are observed between \emph{uncased} and \emph{deacc} models, but both perform consistently above the cased model.


\begin{table}[h]
    \centering
    \begin{tabular}{lccc}
        \hline
        Model          & Hate     &    Sentiment &  Emotion  \\
        \hline
        \bertweet{}    & $\mbf{55.3}$   &      $\mbf{70.3}$ &  $42.8$  \\
        \robertuito{}  & $54.2$   &      $68.4$ &  $44.1$  \\
        \roberta{}     & $45.8$   &      $69.5$ &  $\mbf{46.3}$  \\
        \bert{}        & $48.9$   &      $68.9$ &  $42.8$  \\
        \mbert{}$^*$   & $43.3$   &      $66.6$ &  $40.4$  \\
        \xlmbase{}$^*$ & $45.7$   &      $68.0$ &  $35.7$  \\
        \hline
    \end{tabular}
    \caption{Evaluation results for the three English classification tasks. Results are expressed as the mean Macro F1 score of 10 runs of the classification experiments. $^*$ marks multilingual models}
    \label{tab:results_english}
\end{table}

Table \ref{tab:results_english} displays the results for the evaluation of the selected models for the three tasks in English. We can observe that \robertuito{} outperforms both \mbert{} and \xlm{}, which are the other multilingual models evaluated for the tasks. Compared to monolingual English models, the results of \robertuito{} are similars to those of \roberta{} and slightly above \bert{}. As expected, \bertweet{} obtains the best results.

As for the code-switching evaluation, Table \ref{tab:lince_dev_results} displays the results on the dev dataset of LinCE for NER, POS Tagging, and sentiment analysis. We compare \robertuito{} against other multilingual models such as \mbert{} and \xlmbase{}, and also listing the dev results reported by \cite{winata-etal-2021-multilingual}. The results for \xlm{} and \mbert{} are consistent with the results in that work. A minor improvement is observed but this could an artifact of different choices of hyperparameters or even a slightly different preprocessing.

Finally, Table \ref{tab:lince_benchmark} displays the results from the leaderboard of the \lince{} benchmark\footnote{\url{https://ritual.uh.edu/lince/leaderboard}} for the three selected tasks: Spanish-English sentiment analysis, NER and POS tagging. For sentiment analysis, it obtains the best results in terms of Micro F1. For the other two tasks, it obtains the second position, for which an \xlmlarge{} model \cite{winata-etal-2021-multilingual} has the top results. Among the compared models, \robertuito{} has 108 million parameters, while \xlmlarge{} sums up to around five times this number, making our model the most efficient in terms of size for this subsection of the benchmark (see Appendix \ref{app:model_comparison} for further details on the size of the different models).

\begin{table}[]
    \centering
    \begin{tabular}{lccc}
        \hline
        Model                     & Sentiment       & NER              & POS  \\
        \hline
        \robertuitounc{}          & $\mbf{53.2}$    & $67.2$           & $97.0$ \\
        \robertuitodea{}          & $52.7$          &  $\mbf{67.4}$  & $96.8$ \\
        \robertuitocas{}          & $50.1$          & $66.3$           &  $\mbf{97.3}$ \\
        \mbert{}                  & $51.3$          & $64.8$           & $96.7$ \\
        \xlmbase{}                & $48.8$          & $63.7$           &  $\mbf{97.3}$ \\
        \mbert{}$\dagger$         & --              & $63.7$           &  $\mbf{97.3}$ \\
        \xlmbase{}$\dagger$       & --              & $62.8$           & $97.1$ \\
        \hline
    \end{tabular}
    \caption{Development results for the code-mixed tasks from the LinCE dataset. Sentiment Analysis task performance is measured with Macro F1, Named Entity Recognition with Micro F1, and Part-of-Speech through accuracy. $_U$, $_C$ and $_D$Each figure is the mean of 5 independent runs of the classification experiments. Results marked with $\dagger$ are reported in \protect\newcite{winata-etal-2021-multilingual}}
    \label{tab:lince_dev_results}
\end{table}

\begin{table}[]
    \centering
    \begin{tabular}{lcccc}
        \hline
        Model                    & Sentiment       & NER                 & POS  \\
        \hline
        \robertuito{}            & $\mbf{60.6}$    & $68.5$           & $97.2$     \\
        \xlmlarge{}              & --              & $\mbf{69.5}$     & $\mbf{97.2}$    \\
        \xlmbase{}               & --              & $64.9$           & $97.0$     \\
        C2S \mbert{}    & $59.1$          & $64.6$           & $96.9$     \\
        \mbert{}                 & $56.4$          & $64.0$           & $97.1$     \\
        \bert{}                  & $58.4$          & $61.1$           & $96.9$     \\
        \beto{}                  & $56.5$          & --               & --     \\
        \hline
    \end{tabular}
    \caption{Test results for the code-mixed tasks from Spanish-English section of the \lince{} benchmark. Results are taken from the official leaderboard of this benchmark. Sentiment Analysis performance is measured by Macro F1 score, Named Entity Recognition (NER) with Micro F1 score, and Part-of-Speech (POS) with accuracy. C2S is an acronym for Char2Subword BERT, presented in \protect\cite{aguilar-etal-2021-char2subword-extending}}
    \label{tab:lince_benchmark}
\end{table}

\subsection{Vocabulary efficiency}

\begin{figure}[t]
    \centering
    \includegraphics[width=\columnwidth]{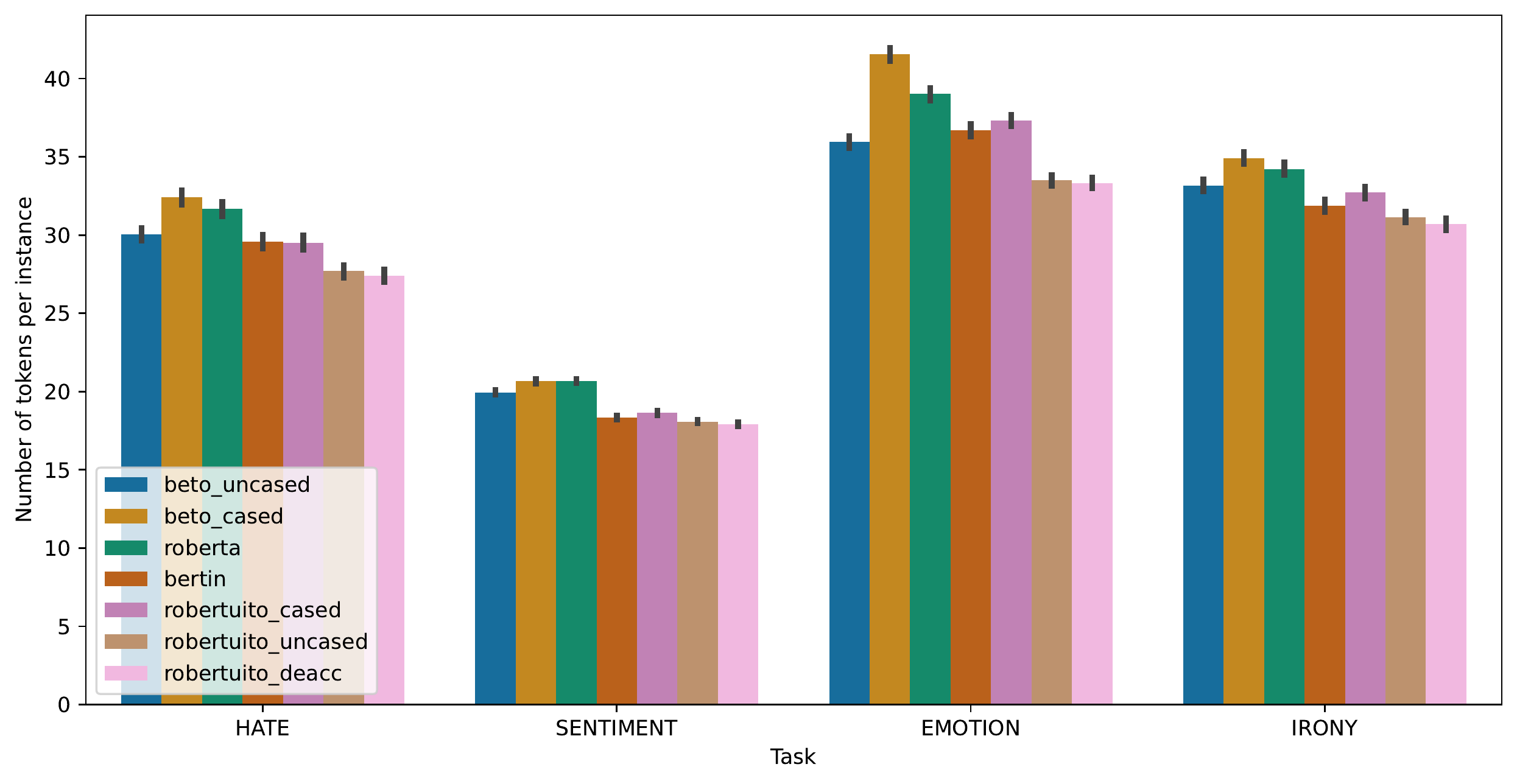}
    \caption{Distribution of the number of tokens per instance. Bars are grouped by task, and display the mean number of tokens per instance with their 95\% confidence interval. Less is better}
    \label{fig:length_tokens}
\end{figure}

Figure \ref{fig:length_tokens} shows the distribution of the number of tokens in the input text for the Spanish tasks. We can observe that \robertuito{} models have more compact representations than \beto{} and \emph{RoBERTa-BNE} for this domain, and, between them, the \emph{deacc} version has a slightly lower mean-length compared to the \emph{uncased} version. \robertuito{} has also shorter representations for the code-mixed datasets, but longer in the case of the English tasks. Appendix \ref{app:vocabulary_efficiency} lists the complete figures for the three evaluation settings.

\section{Discussion}
For this small set of Spanish tasks in the social-media domain, the results show that \robertuito{} outperforms other pre-trained models for Spanish, namely \beto{}, \roberta{} and BERTin. This result is in line with other works which show the effectiveness of social-media-specific language models. A limitation in the evaluation of our model for Spanish tasks is the lack of datasets for other tasks rather than text classification. As far as we know, there are no Spanish-only datasets for sequence labeling (NER and POS tagging) in the social domain.

Among the three proposed variants of \robertuito{}, the cased version is behind the others in terms of performance for most tasks --with the exception of POS tagging and NER-- while the other two (uncased and uncased and deaccented) have comparable performances across all the tasks. We can read this in two ways: one, that a stronger normalization of the input text in Spanish results in no significant improvement in the performance of the model, and two, that keeping accent marks in the input text is neither beneficial nor harmful for the performance of the model.

The reasons for the differences in performance for the uncased models need further investigation. We have some working hypotheses for these differences. First, we believe that accents and non-ASCII characters --with the exception of emojis-- are used in a much more inconsistent way in user-generated text than in more normative text. Therefore, no regularities can be inferred from the data concerning those marks. Second, a bigger amount of data is required to account for the possible differences in meaning for upper case or lower case forms or the lack of difference between them. Future experiments will delve into those two.

Our data collection process for the pre-training stage was centered in Spanish but it allowed other languages and other regional variants to be part of our dataset as well. This point made our model develop some multilingual features, in particular in the code-switching \lince{} benchmark. The results for this benchmark highlights that \robertuito{} is suited for Spanish-English code-mixing tasks, obtaining better results than \mbert{} and matching those of \xlm{}. This comparison, however, is not completely fair because \xlm{} and \roberta{} can handle over one hundred languages but this is not the case for our model. Lastly, the results for the English tasks show that \robertuito{} keeps competitive against monolingual models for the social domain.

\section{Conclusion}
In this work, we presented \robertuito{}, a large-scale model trained on user-generated tweets. We set up a benchmark of classification tasks in social-media text for Spanish, and we showed that \robertuito{} outperforms other available general domain pre-trained language models. Moreover, our model features good code-switching performance in Spanish-English tasks and is competitive against monolingual English models in the social domain.

We proposed three versions of this new model: cased, uncased, and deaccented. We observed that the uncased model performs slightly better than the cased one, and similarly to the deaccented version. Further research is needed to systematize the reasons behind these results.

We have made our pretrained language models public through the \emph{HuggingFace} model hub, and our code can be found at GitHub \footnote{\url{https://github.com/pysentimiento/robertuito}}. We will also make the training corpus available, thus facilitating the development of other models for user-generated Spanish, like word embeddings or other language models. It is even feasible to extract subsets of the corpus representing subdomains of interest, like regional variants of Spanish or specific topics, to develop even more specific models.

Future work includes enhancing our benchmark to include assessment of the performance of such models in open-ended tasks, and experiments for specific subdomains of Spanish.

\section{Acknowledgements}

The authors would like to thank the Google TPU Research Cloud program\footnote{\url{https://sites.research.google/trc/}} for providing free access to TPUs for our experiments. We also thank Juan Carlos Giudici for his help in setting up the GCP environment, and Huggingface's team for their support in the development of RoBERTuito.
\section{Bibliographical References}\label{reference}

\bibliography{robertuito}
\bibliographystyle{lrec2022-bib}

\section{Language Resource References}
\label{lr:ref}
\bibliographystylelanguageresource{lrec2022-bib}
\bibliographylanguageresource{languageresource}

\section*{Appendix}
\label{sec:appendix}

\subsection*{Hyperparameters and training}
\label{app:hyperparameters}

\begin{table}[t]
    \centering
    \small
    \begin{tabular}{ll}
        Hyperparameter  & Value \\
        \hline
        \#Heads           & $12$           \\
        \#Layers          & $12$           \\
        Hidden Size       & $768$          \\
        Intermediate Size & $3072$         \\
        Hidden activation & GeLU           \\
        Vocab. size       & $30,000$         \\
        \hline
        MLM probability   & $0.15  $       \\
        Max Seq length    & $128 $         \\
        Batch Size        & $4,096 $       \\
        Learning Rate     & $3.5 * 10^{-4}$\\
        Decay             & $0.1$          \\
        $\beta_1$         & 0.9            \\
        $\beta_2$         & 0.98           \\
        $\epsilon$        & $10^{-6}$      \\
        Warmup steps      & 36,000 (6\%)   \\
        Training steps    & $600,000$      \\
        \hline
    \end{tabular}
    \caption{Hyperparameters of the \robertuito{} model pre-training.}
    \label{tab:robertuito_architecture}
\end{table}

Table \ref{tab:robertuito_architecture} displays the hyperparameters of the \robertuito{} training for its three versions. For the first prototype of the model, a larger learning rate was tried but it usually diverged near its peak value, so we decided to lower it for the definitive training. Table \ref{tab:training_results} displays the results of the training in terms of cross-entropy loss and perplexity for the three versions of \robertuito{}.

\begin{table}[]
    \centering
    \small
    \begin{tabular}{ll l l}
        Model   & Train loss & Eval loss   & Eval ppl \\
        \hline
        Cased   & 1.864      & 1.753       & 5.772    \\
        Uncased & 1.940      & 1.834       & 6.259    \\
        Deacc   & 1.951      & 1.826       & 6.209 \\
        \hline
    \end{tabular}
    \caption{Training results for each of the three configurations of \robertuito{}, expressed in cross-entropy loss for the Masked-Language-Modeling task and perplexity (ppl)}
    \label{tab:training_results}
\end{table}

\subsection*{Model comparison}
\label{app:model_comparison}
\begin{table}[ht]
    \centering
    \footnotesize
    \resizebox{0.43\textwidth}{!}{
    \begin{tabular}{l c c c }
        \hline
        Model                  & Language                     & Size       & Vocabulary      \\
        \hline
        \robertuito{}          & \multirow{4}{*}{es}          & 108M       & 30K             \\
        \beto{}                &                              & 108M       & 30K             \\
        BERTin                 &                              & 124M       & 50K             \\
        \roberta{}-BNE         &                              & 124M       & 30K             \\

        \hline

        \bert{}                &\multirow{3}{*}{en}           & 109M       & 30K             \\
        \roberta{}             &                              & 124M       & 50K             \\
        \bertweet{}            &                              & 134M       & 64K             \\

        \hline
        \mbert{}               & \multirow{5}{*}{multi}       & 177M       & 105K             \\
        C2S \mbert{}           &                              & 136M       & --            \\
        \xlmbase{}             &                              & 278M       & 250K            \\
        \xlmlarge{}            &                              & 565M       & 250K            \\
        \hline
    \end{tabular}
    }
    \caption{Comparison in terms of model size (measured in number of parameters) and vocabulary sizes for the considered models in this work. C2S \mbert{} is an abbreviation for Char2subword \mbert{} \protect\cite{aguilar-etal-2021-char2subword-extending}}
    \label{tab:model_comparison}
\end{table}

A comparison of the models in terms of the number of parameters and vocabulary sizes is presented in Table \ref{tab:model_comparison}. \robertuito{} is the smallest model in terms of vocabulary size, and as most models share a base architecture (see Table \ref{tab:robertuito_architecture}), this accounts for the difference in the number of parameters.

\subsection*{Vocabulary efficiency}
\label{app:vocabulary_efficiency}

\begin{table}[t]
    \centering
    \textbf{Spanish}
    \footnotesize
    \resizebox{0.45\textwidth}{!}{
    \begin{tabular}{l cccc}
        \hline
        Model                &  EMOTION             &  HATE                 & SENTIMENT            &  IRONY                \\
        \hline
        \robertuito{}$_D$     &$\mbf{33.31}$&$\mbf{27.39 }$ &$\mbf{17.92 }$ &$\mbf{30.70}$ \\
        \robertuito{}$_U$     &     $33.50$    &  $27.70 $     &  $18.08 $     &  $31.12$     \\
        \robertuito{}$_C$    &     $37.33$    &  $29.51 $     &  $18.64 $     &  $32.71$     \\
        \beto{}              &     $35.94$    &  $30.06 $     &  $19.95 $     &  $33.14$     \\
        BERTin               &     $36.68$    &  $29.56 $     &  $18.33 $     &  $31.87$     \\
        \roberta{}           &     $39.02$    &  $31.67 $     &  $20.68 $     &  $34.21$     \\
        \hline
    \end{tabular}
    }

    \footnotesize
    \vspace{1em}
    \textbf{English}

    \resizebox{0.45\textwidth}{!}{
    \begin{tabular}{l ccc}
        \hline
        Model                  &      EMOTION        &          HATE         &    SENTIMENT \\
        \hline
        \bertweet{}            &$\mbf{33.22 }$&$\mbf{29.24 }$ &$\mbf{25.70 }$ \\
        \bert{}                &     $35.20 $     &  $32.05$     &  $26.93 $ \\
        \roberta{}             &     $37.14 $     &  $31.29$     &  $27.00 $ \\
        \robertuito{}$_D$       &     $38.79 $     &  $36.20$     &  $28.82 $ \\
        \robertuito{}$_U$       &     $39.00 $     &  $36.29$     &  $29.22 $ \\
        \robertuito{}$_C$      &     $44.27 $     &  $39.47$     &  $31.49 $ \\
        \hline
    \end{tabular}
    }
    \vspace{1em}
    \footnotesize
    \textbf{Spanish-English Code-switching}

    \resizebox{0.45\textwidth}{!}{
    \begin{tabular}{l c ccc}
        \hline
        Model              &           NER        &          POS      &     SENTIMENT \\
        \hline
        \robertuito{}$_D$   &$\mbf{14.27}$        &$\mbf{8.8}$         &$\mbf{20.65 }$ \\
        \robertuito{}$_U$   &  $14.34$          &  $ 8.84 $          & $20.75    $ \\
        \robertuito{}$_C$  &  $15.41$          &  $ 9.01 $          & $22.14    $ \\
        \beto{}            &  $16.99$          &  $10.78 $          & $24.16    $ \\
        \bert{}            &  $20.88$          &  $10.32 $          & $29.87    $ \\
        \bertweet{}        &  $18.47$          &  $ 9.55 $          & $26.29    $ \\
        \mbert{}           &  $16.47$          &  $ 9.63 $          & $23.82    $ \\
        \xlmbase{}         &  $18.23$          &  $ 9.92 $          & $25.36    $ \\
        \hline
    \end{tabular}
    }

    \caption{Distribution of sentence length measured by number of tokens for each evaluation setting. D, U, and C correspond to deaccented, uncased and cased versions. Results are displayed as the mean of the number of tokens per each sentence. Bold marks best results (less is better)}
    \label{tab:spanish_token_distribution}
\end{table}

Table \ref{tab:spanish_token_distribution} displays the mean sentence length for each considered model and group of tasks. For the Spanish and the code-mixed Spanish-English benchmark, \robertuito{} achieves the more compact representations in mean length in their uncased and deaccented forms. In the case of English, \bertweet{} achieves the shortest representations, with \robertuito{} having longer sequences of tokens than its monolingual counterparts.

\end{document}